\def\year{2019}\relax
\documentclass[letterpaper]{article} 
\usepackage{aaai19}  
\usepackage{times}  
\usepackage{helvet}  
\usepackage{courier}  
\usepackage{url}  
\usepackage{graphicx}  
\usepackage{epsfig}
\begin{document}
%
\title{QADiver: Interactive Framework for Diagnosing QA Models}

\author{Gyeongbok Lee$^{1,2}$\thanks{This work was done during an internship at Clova AI Research, NAVER.}
\hspace{0.7in}Sungdong Kim$^{2}$
\hspace{0.7in}Seung-won Hwang$^{1}$\\
alias\_n@yonsei.ac.kr
\hspace{0.42in}sungdong.kim@navercorp.com
\hspace{0.42in}seungwonh@yonsei.ac.kr\\
$^{1}$Yonsei University\hspace{0.8in} $^{2}$Clova AI Research, NAVER Corp.\\
}
\maketitle

\begin{abstract}
Question answering (QA) extracting answers from text to the given question in natural language, has been actively studied and existing models have shown a promise of outperforming human performance when trained and evaluated with SQuAD dataset. 
However, such performance may not be replicated in the actual setting, for which we need to diagnose the cause, which is non-trivial due to the complexity of model.
We thus propose a web-based UI that provides how each model contributes to QA performances, by integrating visualization and analysis tools for model explanation.
We expect this framework can help QA model researchers to refine and improve their models.

\end{abstract}

\section{Introduction}
With the help of large-scale question answering datasets and deep learning frameworks released to the public, 
question answering (QA) models have been improved rapidly as community efforts. 
Recently, the reported accuracy of state-of-the-art models has exceeded human performance in SQuAD task, 
where plausible answers are extracted for the given question and context document pair. 
However, it is still non-trivial to reproduce such accuracy reported in the paper in production settings.

This paper proposes a diagnosing tool, for troubleshooting the performance gap. For example, is the performance lower than expected because of the biased training to certain types of questions or texts? Is the model attending to wrong words for question or text understanding? Is the embedding suitable for the given task?
At the same time, there is a concerning observation that these models can be easily perturbed by a simple adversarial example added~\cite{jia2017adversarial}.
A desirable tool may support developers to easily perturb training to identify such vulnerability.

We will demonstrate \textbf{QADiver}, a data-centric diagnosing framework for QA model, with diverse interactive visualization and analysis tools for a full pipeline of the attention-based QA model. The framework is connected with the target model to retrieve answer span prediction, inner-model values (such as attention), and no answer probability for given context and question from the model.

Our framework targets SQuAD 2.0 dataset~\cite{rajpurkar2018know}, a machine reading comprehension benchmark that contains both answerable and unanswerable question for given context. In this task, not only finding plausible answer span in context but predicting question is answerable or not is also crucial. 
By exploring large question-answer instances, we expect developers can make a better diagnosis and find insights, than those made from qualitative observations of a few instances. 
The demonstration video for the framework is available in https://youtu.be/V6c8nls6Qcc.

\section{Key Features}

\begin{figure}
\begin{center}
\epsfig{file=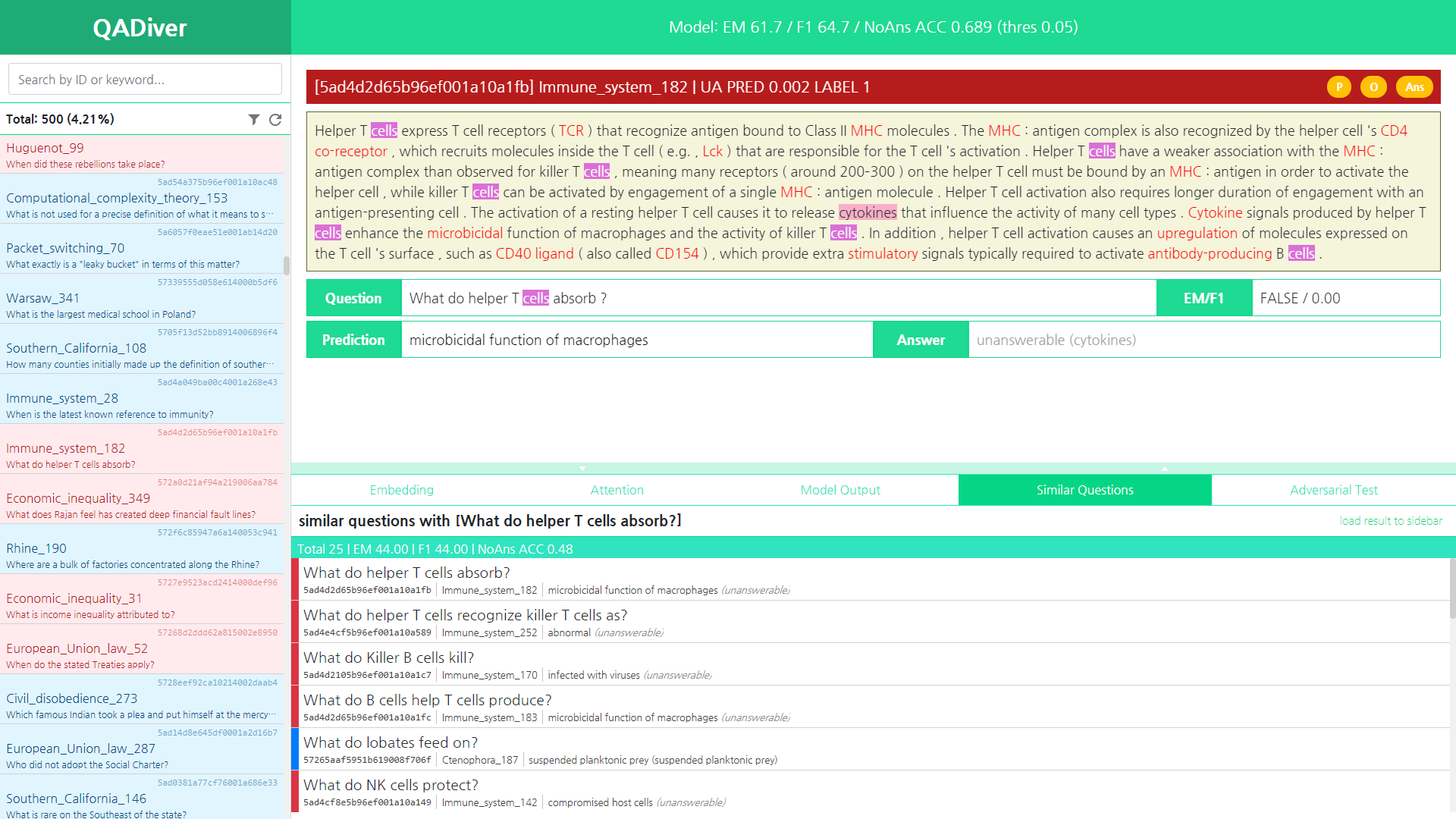, width=80mm}
\caption{Snapshot of QADiver Framework}
\label{fig:snapshot}
\end{center}
\end{figure}

Figure~\ref{fig:snapshot} shows a snapshot of our system, where the cause of low accuracy can be diagnosed as one or more of the following problems:

\subsection{Dataset Bias}

Left-sidebar in Figure~\ref{fig:snapshot} shows data instances form SQuAD 2.0 development set instances.
To support a quick exploration of how such instances are distributed for answers predicted right or wrong, 
we color each to blue and red respectively. This color code is used for the entire system. 
The top right area in Figure~\ref{fig:snapshot} shows more detailed statistics about the instance selected: 
a question, corresponding context, gold answers and prediction result from the model including unanswerable probability and EM/F1 scores. Answer span and out-of-vocabulary words are highlighted. The user can also switch between the original text and preprocessed text by the model for both context and question to identify cases where the underlying NLP models, such as tokenizer, are not working as intended.

\subsection{Embedding Analysis}

QA model performance also depends on the effectiveness of word embedding.
To visualize such effectiveness, we allow developers to visualize word vector (and nearby words in the space by clicking a word.
The user can restrict words in the context, or use the whole vocabulary set for the dataset. For a fast similarity search of dense vectors, we use FAISS~\cite{JDH17} library.

\subsection{Neural Model Internals}

Visual representation for internal states of the neural model shows whether right words are highlighted for understanding questions or finding answers.
The most common example is visualizing attention and output layer in the model. For attention layer, we visualize a context-question attention matrix heatmap for the given instance. 
Similarly, we provide an interpreted version of the model output used for answer span prediction and answerability decision by listing the top-k words with the highest weights. Users can also see the list of answer span candidates (including ``unanswerable" case) and its certainty as a colored heatmap.

\subsection{Question Bias}

A model can be biased to answer a certain type of questions particularly well.
To diagnose such a case, we identify similar questions to the given instance, labeled with prediction result and evaluation metrics (EM/F1). 
For a desirable question embedding, projecting similar question types close in the embedding space, we use well-studied features from the question and gold answer such as: answer length, the existence of number and entity, and 2-word question prefix like \textit{What is} and \textit{How many}.
Feature values of questions in the same class are mean-aggregated to generate a global statistic vector. To represent the characteristic of each question, local features like word match ratio between context and question and one-hot vector for frequent words are used so that similar types of questions have high similarity. As each question is vectorized, top similar questions for the instance can be retrieved from the whole dataset by the similarity search. 

\subsection{Adversarial Test}
As we overviewed, many existing QA models are reportedly lacking the robustness over adversarial examples.
Using our tool, the user can easily perform the adversarial test for each instance in two ways: manual modification and rule-based test. First, the user can modify some words in context document and question from the data viewer by double-clicking target word and replacing to another. After this edit, we show an updated prediction and EM/F1 score.

Instead of perturbing each word, which may be costly, the user may create reusable adversarial rules, about word and its 
part-of-speech (POS) tag. We use NLTK toolkit~\cite{bird2009natural} for POS tagging and word tokenization used in rule matching. We also provide pre-defined adversarial rules from SEAR~\cite{ribeiro2018semantically} for those wanting to check
the robustness for common cases.

\section{Related Work}
Due to the growing demand for model interpretability, many visualization tools for QA models were proposed:
AllenNLP machine comprehension demo\footnote{http://demo.allennlp.org/machine-comprehension} provides answer span highlights and attention matrix visualization.
\cite{rueckle:2017:ACL} shows attention visualization on context and question text, and provides a comparison between the two model.
\cite{shusen2018visual} proposes bipartite graph attention representation and hierarchical visual for highly asymmetric attention. This tool also supports word- and attention-level perturbation by user edit.

\section{Acknowledgements}
This work was supported by the ICT R\&D program of MSIT/IITP. [No.2017-0-01778, Development of Explainable Human-level Deep Machine Learning Inference Framework]

\bibliography{aaai19}

\begin{thebibliography}{}

\bibitem[\protect\citeauthoryear{Bird, Klein, and
  Loper}{2009}]{bird2009natural}
Bird, S.; Klein, E.; and Loper, E.
\newblock 2009.
\newblock {\em Natural language processing with Python: analyzing text with the
  natural language toolkit}.
\newblock " O'Reilly Media, Inc.".

\bibitem[\protect\citeauthoryear{Jia and Liang}{2017}]{jia2017adversarial}
Jia, R., and Liang, P.
\newblock 2017.
\newblock Adversarial examples for evaluating reading comprehension systems.
\newblock In {\em Proceedings of the 2017 Conference on EMNLP}.

\bibitem[\protect\citeauthoryear{Johnson, Douze, and J{\'e}gou}{2017}]{JDH17}
Johnson, J.; Douze, M.; and J{\'e}gou, H.
\newblock 2017.
\newblock Billion-scale similarity search with gpus.
\newblock {\em arXiv preprint arXiv:1702.08734}.

\bibitem[\protect\citeauthoryear{Rajpurkar, Jia, and
  Liang}{2018}]{rajpurkar2018know}
Rajpurkar, P.; Jia, R.; and Liang, P.
\newblock 2018.
\newblock Know what you don't know: Unanswerable questions for squad.
\newblock {\em Proceedings of the 56th Annual Meeting of the ACL}.

\bibitem[\protect\citeauthoryear{Ribeiro, Singh, and
  Guestrin}{2018}]{ribeiro2018semantically}
Ribeiro, M.~T.; Singh, S.; and Guestrin, C.
\newblock 2018.
\newblock Semantically equivalent adversarial rules for debugging nlp models.
\newblock In {\em Proceedings of the 56th Annual Meeting of the ACL}, volume~1,
   856--865.

\bibitem[\protect\citeauthoryear{R{\"u}ckl{\'e} and
  Gurevych}{2017}]{rueckle:2017:ACL}
R{\"u}ckl{\'e}, A., and Gurevych, I.
\newblock 2017.
\newblock End-to-end non-factoid question answering with an interactive
  visualization of neural attention weights.
\newblock In {\em Proceedings of the 55th Annual Meeting of the ACL-System
  Demonstrations (ACL 2017)},  19--24.
\newblock Association for Computational Linguistics.

\bibitem[\protect\citeauthoryear{Shusen~Liu and
  Bremer}{2018}]{shusen2018visual}
Shusen~Liu, Tao~Li, Z. L. V. S. V.~P., and Bremer, P.-T.
\newblock 2018.
\newblock Visual interrogation of attention-based models for natural language
  inference and machine comprehension.
\newblock In {\em Proceedings of the 2018 Conference on EMNLP: System
  Demonstrations}.

\end{thebibliography}
\bibliographystyle{aaai}

\end{document}